\documentclass[runningheads]{llncs}
\usepackage{graphicx}
\usepackage{algpseudocode}
\usepackage{algorithm}
\usepackage{booktabs}
\usepackage{subcaption}
\usepackage{multicol}
\usepackage{bm}
\usepackage{amsfonts,mathtools}
\newcommand{\vv}[1]{\mathbf{#1}}
\newcommand\defas{\mathrel{\overset{\makebox[0pt]{\mbox{\normalfont\tiny\sffamily def}}}{=}}}

\begin{document}
\title{Active and sparse methods in smoothed model checking}
\author{Paul Piho, Jane Hillston}
\institute{University of Edinburgh}

\maketitle

\begin{abstract}
    Smoothed model checking based on Gaussian process classification provides a
    powerful approach for statistical model checking of parametric continuous
    time Markov chain models.
    The method constructs a model for the functional dependence of satisfaction
    probability on the Markov chain parameters. 
    This is done via Gaussian process inference methods from a limited
    number of observations for different parameter combinations.
    In this work we consider extensions to smoothed model checking based on
    sparse variational methods and active learning.
    Both are used successfully to improve the scalability of smoothed model checking.
    In particular, we see that active learning-based ideas for iteratively
    querying the simulation model for observations can be used to steer the
    model-checking to more informative areas of the parameter space and thus
    improve sample efficiency. 
    Online extensions of sparse variational Gaussian process inference
    algorithms are demonstrated to provide a scalable method for implementing active
    learning approaches for smoothed model checking.
\end{abstract}

\section{Introduction}
Stochastic modelling coupled with verification of logical properties via
model checking has provided useful insights into the behaviour of the stochastic
models from epidemiology, systems biology and networked computer systems.
A large number of interesting models in these fields are too complex for the
application of exact model checking methods~\cite{KwiatkowskaNP07}.
To improve scalability of model checking there has been significant work on
statistical model checking that aims to estimate the satisfaction probability
of logical properties based on independently sampled trajectories of the
stochastic model~\cite{AghaP18}. 

This paper considers statistical model checking in the context of parametrised
continuous time Markov chain models. 
Statistical model checking methods have generally considered single
parametrisations of a model.
Based on a large number of independent sample trajectories one can estimate the
probability of the model satisfying a specified logical property defined over
individual sample trajectories.
In order to gain insight across the entire parameter space associated with a
model, it can be necessary to repeat the estimation procedures with different 
parametrisations to cover the whole space, which leads to poor scalability.

As an alternative, a model checking approach based on Gaussian process
classification, named smoothed model checking, was proposed in
~\cite{BortolussiMS16}.
The main result of that paper was to show that under mild conditions the
function mapping parameter values to satisfaction probabilities is smooth.
Thus the problem can be solved as a Gaussian process classification problem
where the aim is to estimate the function describing the satisfaction
probability over the parameter space.
Model checking results, returning a label true or false, of individual
simulation trajectories are used as the training data to infer how the
satisfaction probability depends on the model parameters. 
This method can be used to greatly reduce the number of simulation trajectories
needed to estimate the satisfaction probability in exchange for some accuracy. 

There are two aspects that limit the speed of such model checking procedures.   
Firstly, the computational cost of gathering the individual trajectories and
secondly, the cost of (approximate) Gaussian process inference itself. 
Both benefit from keeping the number of gathered trajectories as low as possible 
while minimising the impact a smaller set of training data has on the
accuracy of the methods.
In order to keep the gathered sample size small we propose a method based on
active learning.
In particular, we make the observation that the parameter space of models is
usually constrained to physically reasonable ranges.
However, even when constrained to such ranges there can be large parts of the
parameter space where the probability of satisfying a formula exhibits stiff
behaviour.
Adaptively identifying stiff and non-stiff parts of the parameter space in order
to decide where to concentrate the computational effort leads to improved
algorithms for smoothed model checking.

Our approach is based on moving the smoothed model checking approach in the
context of continuous time Markov chains to an online setting.
This allows us to implement effective active sampling strategies for the
parameter space of the model that take into account the already gathered
information. 
In order to make this approach scalable we leverage and demonstrate the usage
of state of the art sparse variational Gaussian process inference methods.
In particular, we consider streaming variational inference with inducing
points~\cite{Bui2017}.

\section{Related work}
A wealth of literature exists on statistical model checking of stochastic
systems.
The use of statistical methods in the domain of formal verification is
motivated by the fact that in order to perform statistical model checking it is
only necessary to be able to simulate the model.
Thus these methods can be used for systems where exact verification methods are
infeasible including black-box systems~\cite{Legay2019}.
In its classical formulation, this involves hypothesis testing~\cite{Sen2004}
with respect to the desired (or undesired) property based on independent trials,
or in this case, stochastic simulations.

In addition to the frequentist approaches based on hypothesis testing, there have
been Bayesian approaches~\cite{Jha2009} to estimate the satisfaction probability
of a given logical formula.
Our work follows the approach presented in~\cite{BortolussiMS16} where the
dependence of the satisfaction probability on model parameters is modelled as a
Gaussian process classification problem.

The problem of deciding where to concentrate the model checking efforts is
closely related to optimal experimental design.
Experimental design problems are commonly treated as optimisation problems where
the goal is to allocate resources in a way that allows the experimental goals to
be reached more rapidly and thus with smaller costs~\cite{Santner2003}.
This idea is also known in the machine learning literature as active
learning~\cite{Settles2012}. 
The idea is to design learning algorithms that interactively query an oracle to
label new data points.


In the context of model checking, active learning was used
in~\cite{BortolussiS18} to solve a closely related threshold synthesis problem.
That approach used a base grid on the parameter space for initial estimation.
The estimates were then refined around values where the satisfaction probability
was close to a defined threshold.
However, the threshold for synthesis has to be defined a priori making the
introduced active step not applicable when we are interested in the
satisfaction probability.
We further address the scalability of the ideas presented by the authors
of~\cite{BortolussiS18} by considering sparse approximation results for Gaussian.

\section{Background}

\subsection{Continuous time Markov chains}
\label{sec:ctmc}
Stochastic models are widely used to model a variety of phenomena in natural and
engineered systems.
We focus on a type of stochastic model commonly used in biological modelling,
epidemiology and performance evaluation domains.
Specifically, we consider continuous time Markov chain models (CTMCs).
To define a CTMC we start by noting that it is a continuous-time stochastic
process and thus defined as an indexed collection of random variables
$\{\vv{X}\}_{t \in \mathbb{R}_{\geq 0}}$.
We consider CTMCs defined over a finite state space $S$ with an $|S| \times
|S|$ matrix $Q$ whose entries $q(i, j)$ satisfy
\begin{multicols}{3}
    \begin{enumerate}
        \item $0 \leq -q(i,i) < \infty$
        \item $0 \leq q(i,j)$ for $i \neq j$
        \item $\sum_{j}q(i,j) = 0$  
    \end{enumerate}
\end{multicols}
A CTMC is then defined by the following: for time indices $t_1 <
t_2\ldots < t_{n+1} $ and states $i_1, i_2, \ldots, i_{n+1}$ we have 
\begin{equation*}
    \mathbb{P}(\vv{X}_{t_{n+1}} = i_{n+1} | \vv{X}_{t_n} = i_n, \ldots, \vv{X}_{t_1} = i_1) = p(i_{n+1}; t_{n+1} | i_n; t_n)
\end{equation*}
where $p(j; t | i; s)$ is the solution to the following Kolmogorov forward equation
\begin{equation*} 
    \frac{\partial}{\partial t}p(j;t | i; s) = \sum_{k}p(k;t
    | i; s)q(k, j), \qquad \mbox{on $(s, \infty)$ with $p(j; s | i;s) =
    \delta_{ij}$} 
\end{equation*}
with $\delta_{ij}$ being the Kronecker delta taking the value $1$ if $i$ and
$j$ are equal and the value $0$ otherwise.
By convention the sample trajectories of CTMCs are taken to be right-continuous.

In the rest of the paper we consider parametrised models $\mathcal{M}_{\vv{x}}$
and assume that the model $\mathcal{M}$ for a fixed parametrisation $\vv{x} \in
\mathbb{R}^k$ defines a CTMC.
Thus, the model $\mathcal{M}$ specifies a function mapping parameters $\vv{x}$
to generator matrix $Q$ of the underlying CTMC.
A commonly studied special class of CTMC models are population CTMCs where each
state of the CTMC corresponds to a vector of counts.
These counts are used to model the aggregate counts of groups of
indistinguishable agents in a system.
In biological modelling and epidemiology such models are often defined as
chemical reaction networks (CRN).

\begin{example} 
    Let us consider the following SIR model defined as a CRN
    \begin{align*} 
        S + I & \xrightarrow{k_I} I + I \qquad\qquad I \xrightarrow{k_R} R
    \end{align*} 
    where $S$ gives the number of susceptible, $I$ the infected and
    $R$ the recovered individuals in the system.
    The first type of transition corresponds to an infected and
    susceptible individuals interacting, resulting in the number of infected
    individuals increasing and the number of susceptible decreasing.
    The second type of transition corresponds to recovery of an infected
    individual and results in the number of infected decreasing and the number of
    recovered increasing. 
    The states of the underlying CTMC keep track of the counts of different
    individuals in the system. 
    For the example let us set the initial conditions to $(95, 5, 0)$ ---  at
    time $0$ there are $95$ susceptible, $5$ infected and $0$ recovered
    individuals in the system.
    The parameters $k_I$ and $k_R$ give the infection and recovery rates
    respectively.  
    We revisit this example throughout the paper to illustrate the presented
    concepts.
\end{example} 

\subsection{Smoothed model checking}
\label{sec:model checking}
Smoothed model checking was introduced in~\cite{BortolussiMS16} as a scalable
method for statistical model checking where Gaussian process classification
methods were used to infer the functional dependence between a parametrisation
of a model and the satisfaction probability given a logical specification.

As described in Section~\ref{sec:ctmc}, suppose we have a model
$\mathcal{M}_{\vv{x}}$ parametrised by vector of values $\vv{x} \in
\mathbb{R}^d$ such that the model $\mathcal{M}$ for a fixed parametrisation
$\vv{x}$ defines a CTMC.
Additionally assume we have a logical property $\varphi$ we want to check against.
The logical properties we consider here are defined as a mapping from the
time trajectories over the states of $\mathcal{M}_{\vv{x}}$ to $\{0, 1\}$
corresponding to whether the property holds for a given sample trajectory of
$\mathcal{M}_{\vv{x}}$ or not.
One way to define such mappings would be, for example, to specify the properties
in metric interval temporal logic (MiTL)~\cite{MalerN04} or signal temporal
logic (STL)~\cite{DonzeM10} and map the paths satisfying the properties to $1$
and those not satisfying the properties to $0$.
Through sampling multiple trajectories for the same parametrisation we gain an
estimate of the satisfaction probability corresponding to the parametrisation.

With that in mind, a logical property $\varphi$ with respect to
$\mathcal{M}_{\vv{x}}$ can be seen to give rise to a Bernoulli random variable.
The binary outcomes of the random variable correspond to whether or not a
randomly sampled trajectory of $\mathcal{M}_{\vv{x}}$ satisfies the property
$\varphi$.
We introduce the notation $f_{\varphi}(\vv{x})$ for the parameter
of the said Bernoulli random variable given the model parameters $\vv{x}$.
In particular, samples from the distribution
$\mathit{Bernoulli}(f_{\varphi}(\vv{x}))$ model whether a randomly sampled
trajectory of $\mathcal{M}_{\vv{x}}$ satisfies $\varphi$ --- for a parameter
value $\vv{x}$ the logical property is said to be satisfied with probability
$f_{\varphi}(\vv{x})$.

A naive approach for estimating $f_{\varphi}(\vv{x})$ at a given parametrisation
$\vv{x}$ is to gather a large number $N$ of sample trajectories and give simple
Monte Carlo estimate for the $f_{\varphi}(\vv{x})$ by dividing the number of
trajectories where the property holds by the total number of sampled
trajectories $N$.
An accurate estimate requires are large number of samples.
However, having such estimate at a set of given parametrisations does no provide
us with a rigorous way to estimate the satisfaction function at a nearby point.

In~\cite{BortolussiMS16} the authors considered population CTMCs.
It was shown that the introduced satisfaction probability $f_{\varphi}(\vv{x})$
is a smooth function of $\vv{x}$ under the following conditions: the transition
rates of the CTMC $\mathcal{M}_{\vv{x}}$ depend smoothly on the parameters
$\vv{x}$; and the transition rates depend polynomially on the state vector
$\vv{X}$ of the CTMC.

The result was exploited by treating the estimation of the satisfaction function
$f_{\varphi}(\vv{x})$ as a Gaussian process classification problem.
The main benefit of this approach is that, based on sampled model checking
results, we can reconstruct an approximation for the functional dependence
between the parameters and satisfaction probability. 
This makes it easy to make predictions about the satisfaction probability at
previously unseen parametrisations.

Simulating $\mathcal{M}_{\vv{x}}$ we gather a finite set of
observations $\mathcal{D} = \{(\vv{x}_i, y_i) | i = 1, \cdots,
n\}$ where $\vv{x}_i$ are the parametrisations of the model and $y_i$ correspond
to model checking output over single trajectories.  
For classification problems, a Gaussian process prior with mean $m$ and kernel
$k$ is placed over a latent function 
\begin{equation*}
    g_{\varphi}(\vv{x}) \sim \mathit{GP}(m(\vv{x}), k(\vv{x}, \vv{x}'))
\end{equation*}
Here, let us consider the standard squared exponential kernel defined by
\begin{equation*}
    k(\vv{x}, \vv{x}') = \exp\left(-\frac{|\vv{x} - \vv{x}'|^2}{2\ell}\right)
\end{equation*}
where $\ell$ is the length scale parameter governing how far two distinct points
have to be in order to be considered uncorrelated.

The function $g_{\varphi}$ is then squashed through the standard logistic or
probit transformation $\sigma$ so that the composition
$\sigma(g_{\varphi}(\vv{x}))$ takes values between 0 and 1.
The quantity $\sigma(g_{\varphi}(\vv{x}))$ is interpreted as the probability
that $\varphi$ holds given model parametrisation $\vv{x}$ and thus estimates
the probability $f_{\varphi}(\vv{x})$ that a simulation trajectory for
parameters $\vv{x}$ satisfies the property $\varphi$. 

The general aim of Gaussian process inference is to find the distribution 
$p(g_{\varphi}(\vv{x}^*) | \mathcal{D})$ over the values $g_{\varphi}$ at some
test point $\vv{x}^*$ given the set of training observations $\mathcal{D}$.
This distribution is then used to produce a probabilistic prediction at
parameter $\vv{x}^*$ of $\sigma(g_{\varphi}(\vv{x}^*)) \approx
f_{\varphi}(\vv{x}^*$).
We present details of inference in the next section.
This section is ended by returning to the running SIR example.

\begin{figure}[t!]
    \centering
    \includegraphics[width=0.60\textwidth]{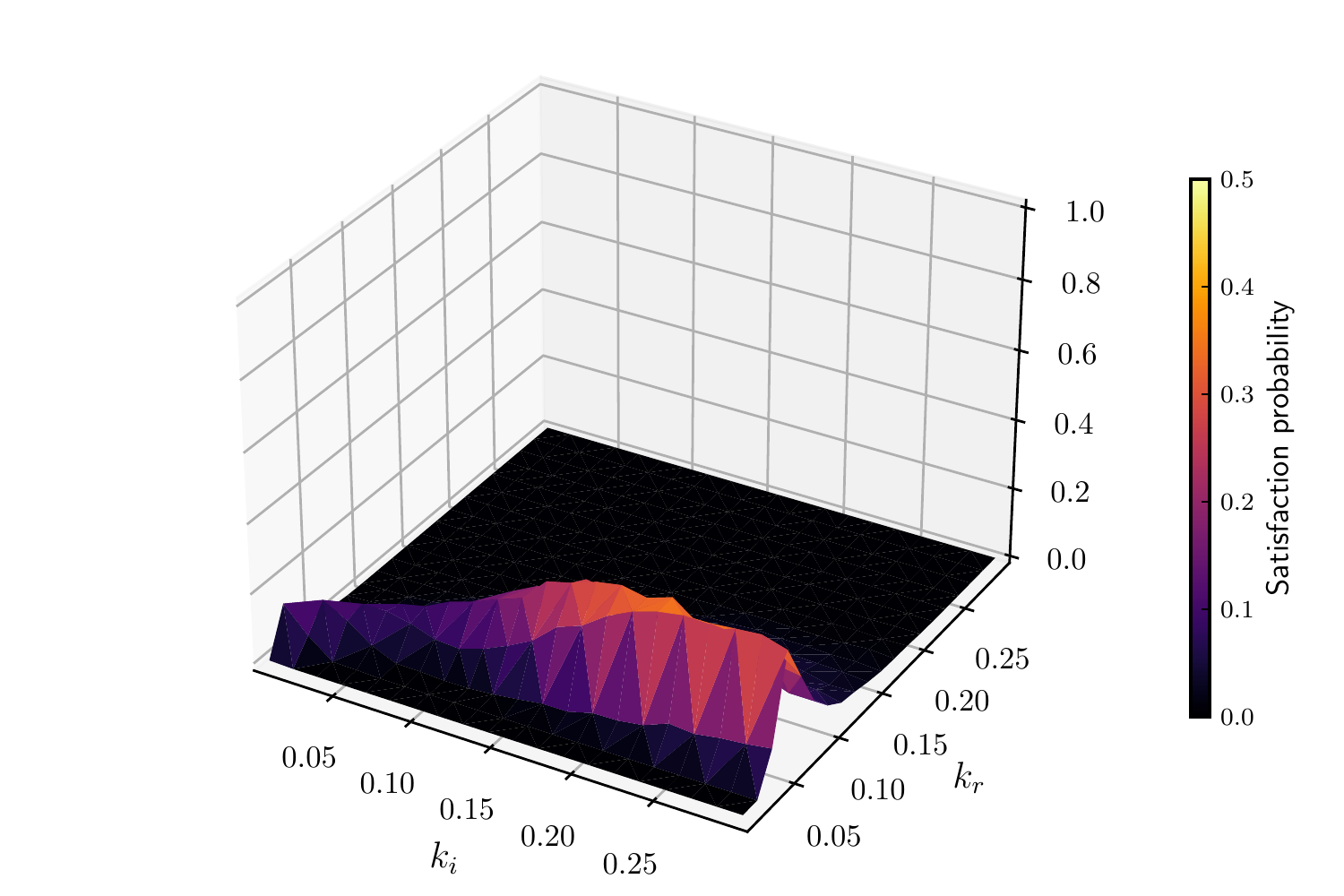}
    \caption{Baseline satisfaction probability surface.}
    \label{fig:baseline surface}
\end{figure}

\begin{figure}[t!]
    \centering
    \includegraphics[width=0.7\textwidth]{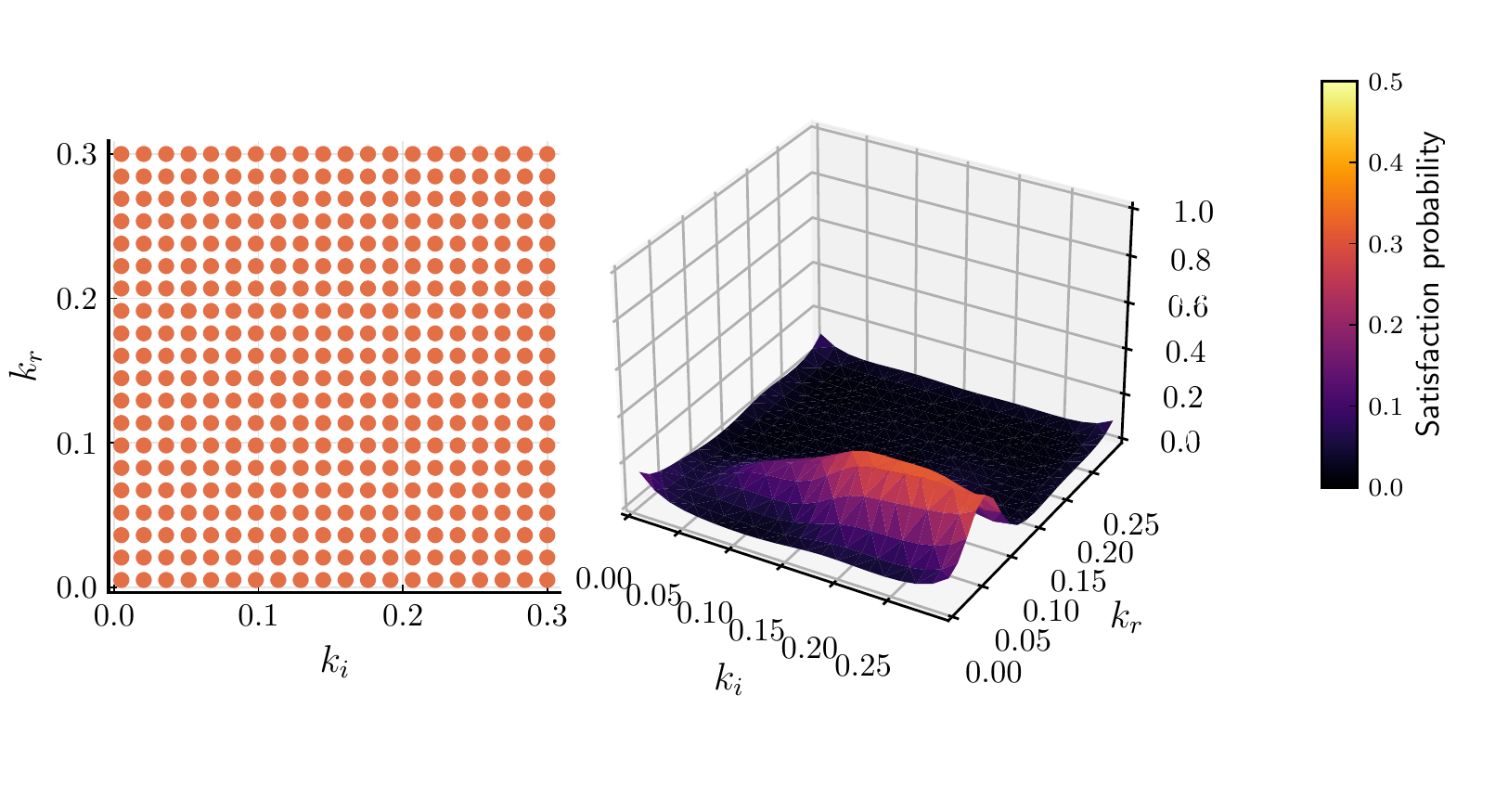}
    \vspace{-2em}
    \caption{Smoothed model checking satisfaction probability surface. $144$ test points.}
    \label{fig:smoothed surface}
\end{figure}

\begin{example} 
    The property we consider is the following: 
    there always exists an infected agent in the population in the time interval
    $(0.0, 100.0)$ and in the time interval $(100.0, 120.0)$ the number of infected
    becomes $0$.
    Constraining the parameters to the ranges $k_I \in [0.005, 0.3]$ and $k_R \in
    [0.005, 0.3]$ gives satisfaction probabilities as depicted in the
    Figure~\ref{fig:baseline surface}.
    There each estimate on the $20$-by-$20$ grid is calculated based on $2000$
    stochastic simulation sample runs of the model.
    For comparison, Figure~\ref{fig:smoothed surface} gives the results of the
    smoothed model checking where $10$ sample trajectories are drawn for each
    parameter on the $12$-by-$12$ grid.
    The smoothed model checking approximation for the model checking problem  
    shows good agreement with the baseline surface and is much faster to perform.
\end{example} 

\subsection{Variational inference with inducing points}
\label{sec:var inference}
In order to infer the latent Gaussian process $g_{\varphi}$ based on training
data $\mathcal{D}$ we have to deal with two problems.
Firstly the inference is analytically intractable due to the non-Gaussian
likelihood model provided by Bernoulli observations.  
To counter this there exist a wealth of approximate inference schemes like Laplace
approximation, expectation propagation~\cite{Minka2001,Hernandez-Lobato2016},
and variational inference methods~\cite{Titsias2009}.   
Here we consider variational inference. 
The second problem is that the methods for inference in Gaussian process models
have cubic complexity in the number of training cases.
To address that there exist sparse approximations based on inducing variables.
Sparse variational methods~\cite{Csato2002,Titsias2009} are popular methods for
reducing the complexity of Gaussian process inference by constructing an
approximation based on a small set of inducing points that are typically
selected from training data.
In this section we detail the inference procedure.

Variational inference methods choose a parametric class of variational
distribution for the posterior and minimising the KL-divergence between the real
posterior and the approximate posterior.
To accommodate large training data sets we work with sparse variational methods.
We start by defining the prior distribution 
\begin{equation*}
    p(\vv{g}_{\varphi}, \vv{g}_\vv{u}) = \mathcal{N}
        \left(
        \begin{bmatrix} \vv{g}_{\varphi} \\ \vv{g}_\vv{u} \end{bmatrix}
            \Bigg| \vv{0},
        \begin{bmatrix} 
            \vv{K}_{nn} & \vv{K}_{nm} \\
            \vv{K}_{nm}^T & \vv{K}_{mm} \\
        \end{bmatrix}
        \right)
\end{equation*}
where $\vv{g}_{\varphi}$ is a vector of $n$ latent function values at
$g_{\varphi}^i = g_{\varphi}(x_i)$ evaluated at $\vv{x} = [\vv{x}_1, \cdots, \vv{x}_n]$.
Similarly, $g_{\vv{u}}^i = g_{\varphi}(\vv{u}_i)$ is a vector of $m$ latent function
values evaluated at chosen \emph{inducing points} $\vv{u} = [\vv{u}_1, \cdots, \vv{u}_m]$.
The matrices $\vv{K}_{nn}$, $\vv{K}_{nm}$ and $\vv{K}_{mm}$ are defined by the
kernel function.
In particular, the $(i,j)$-th element of the matrix $\vv{K}_{nn}$ is given by
$k(\vv{x}_i, \vv{x}_j)$. 
Similarly, $\vv{K}_{nm}$ gives the kernel matrix between the training points
$\vv{x}$ and the inducing points $\vv{u}$ and $\vv{K}_{mm}$ gives the kernel
matrix between the locations of inducing points.
We then fit the variational posterior at those points rather
than the whole set of training data points.
The assumption we are making is that $p(g_{\varphi}(\vv{x}_*) | \vv{g}_{\varphi},
\vv{u}) = p(g_{\varphi}(\vv{x}_*) | \vv{u})$.
That is, the inducing points $\vv{u}$ are a sufficient statistic for a function
value at a test point $\vv{x}_*$.

Under this assumption we make predictions at a test point $\vv{x}^*$ as follows: 
\begin{equation*}
    p(g_{\varphi}(\vv{x}_*) | \vv{y}) 
    = \int p(g_{\varphi}(\vv{x}_*), \vv{u} | \vv{y}) d\vv{u}
    = \int p(g_{\varphi}(\vv{x}_*) | \vv{u}) p(\vv{u} | \vv{y}) d\vv{u}
\end{equation*}
Thus, we need posterior distribution $p(\vv{u}|\vv{y})$ at the inducing
points.
From Bayes rule we know that $p(\vv{u}|\vv{y}) = \frac{p(\vv{y}|\vv{u})
p(\vv{u})}{p(\vv{y})}$.
The terms $p(\vv{y})$ and $p(\vv{y} | \vv{u})$ are called the likelihood and the
marginal likelihood respectively.
The above integral can be seen as the weighted average of the prior model
$p(g_{\varphi}(\vv{x}_*) | \vv{u})$ with weight being determined by the
posterior distribution.
Here, as mentioned, we consider variational approximations where $p(\vv{u} |
\vv{y})$ is approximated by a multivariate Gaussian $q(\vv{u})$ making the
expression for $p(g_{\varphi}(\vv{x}_*) | \vv{y})$ tractable.

Finding the parameters of $q(\vv{u})$ is done by minimising the KL divergence
between $q(\vv{u})$ and $p(\vv{u} | \vv{y})$.
\begin{align}
    \mathcal{D}_{\mathit{KL}}(q(\vv{u}), p(\vv{u}|\vv{y})) 
       & = \int q(\vv{u}) \log \frac{q(\vv{u})}{p(\vv{u} | \vv{y})} d\vv{u}
    = - \big\langle \log \frac{p(\vv{y}, \vv{u})}{q(\vv{u})} \big\rangle_{q(\vv{u})}
    + \log p(\vv{y}) \label{eq:kl variational}
\end{align}
The term $\log p(\vv{y})$ is known as the log marginal likelihood.
In the following we use the well-known Jensen's inequality\footnote{For a
concave function $f$ and a random variable $X$ we have
the following well-known inequality: $f\langle X \rangle \geq \langle f(X)
\rangle$.} to derive a lower lower bound for the log marginal likelihood.
As log function in concave we get the following.
\begin{align*}
    \log p(\vv{y}) & = \log \int p(\vv{y}, \vv{u}) d\vv{u} 
    = \log \Big\langle \frac{p(\vv{y}, \vv{u}) }{q(\vv{u})} \Big\rangle_{q(\vv{u})} 
    \geq \Big\langle \log \frac{p(\vv{y}, \vv{u}) }{q(\vv{u})} \Big\rangle_{q(\vv{u})} \\
    & = \langle \log p(\vv{y}|\vv{u}) \rangle_{q(\vv{u})} - \mathcal{D}_{\mathit{KL}}(q(\vv{u}), p(\vv{u}))
\end{align*}
The last line of the above equation is known as the evidence-based lower bound
or ELBO.  
Now note that the first term in the expression for KL-divergence in
Equation~\ref{eq:kl variational} is exactly the derived ELBO.
As the KL-divergence is always non-negative then maximising the ELBO minimises
the KL-divergence between the approximate and true posteriors $q(\vv{u})$ and
$p(\vv{u} | \vv{y})$.

Choosing our approximating family of variational distribution to be multivariate
Gaussian makes the KL term in ELBO easy to evaluate.
The integral in the expectation term can be computed via numerical approximation
schemes making in possible to use ELBO as a utility function for optimising the
parameters of the approximate posterior $q(\vv{u})$.
When $q(\vv{u})$ is chosen to be a multivariate Gaussian these parameters are
the mean $\bm{\mu}$ and covariance matrix $\bm{\Sigma}$.

The final part of this section briefly discusses making predictions based on the 
approximate posterior $q(\vv{u})$.
With the approximation $p(\vv{u} | \vv{y}) \approx q(\vv{u})$ the predictions
are given by the following integral 
\begin{equation*}
    p(g_{\varphi}(\vv{x}_*) | \vv{y}) 
    \approx \int p(g_{\varphi}(\vv{x}_*) | \vv{u}) q(\vv{u}) d\vv{u}
\end{equation*}
This can be shown~\cite{Rasmussen2006} to be a probability density function of a Normal distribution with the following mean and covariance matrix
\begin{align*}
    &k(\vv{x}_*, \vv{u}) K_{mm}^{-1} \bm{\mu} \\
    &k(\vv{x}_*, \vv{x}_*) - k(\vv{x}_*, \vv{u}) K_{mm}^{-1}\left[ K_{mm} - \bm{\Sigma} \right]^{-1} \left[k(\vv{x}_*, \vv{u}) K_{mm}^{-1}\right]^T
\end{align*}

\subsection{Inducing points}
\label{sec:inducing points}
Previously we introduced the idea of sparse variational inference
where the posterior distribution is fitted to a selection of inducing points
$\vv{u}$ such that $|\vv{u}|$ is much smaller than the size of the whole
training data set.
In this section we discuss how to choose the set of inducing points $\vv{u}$.
A common approach is to use cluster centers from k-means++ clustering as the
inducing points~\cite{David2007}.
In this papers we simply take a regular grid of inducing points.

\subsection{Active learning}
\label{sec:active learning}
Active learning methods in machine learning are a family of methods which may
query data instances to be labelled for training by an
\emph{oracle}~\cite{Settles2011}.
The fundamental question asked by active learning research is whether or not these
methods can achieve higher accuracy than passive methods with fewer labelled
examples.
This is closely related to the established area of optimal experimental design,
where the goal is to allocate experimental resources in a way that reduces
uncertainty about a quantity or function of
interest~\cite{Santner2003,Settles2012}.

In the case of Gaussian process classification problems like smoothed model checking
an active learning procedure can be set up as follows.
An \emph{active learner} consists of a classifier learning algorithm
$\mathcal{A}$ and a \emph{query function} $\mathcal{Q}$.
The query function is used to select an unlabelled sample $u$ from the pool of
unlabelled samples $\mathcal{U}$.
This sample is then labelled by an oracle.
In the case of stochastic model checking the pool of unlabelled samples
$\mathcal{U}$ corresponds to a subset of the possible parametrisations for the
model.  
An oracle is implemented by running the stochastic simulation for the selected
parametrisation and model checking the resulting trajectory.

The above describes a pool-based active learner.
Common formulations of such pool-based learners select a single unlabelled
sample at each iteration to be sampled.
However, in many applications it is more natural to acquire labels for multiple 
training instances at once.
In particular, the query function $\mathcal{Q}$ selects a subset $U \subset
\mathcal{U}$.
We see in the next section that the sparse inference methods can be extended to
a setting where batches of training data become available over time making it
natural to decide on a query function that selects batches of queries.
The main difficulty of selecting a batch of queries instead of a single query 
is that the instances in the subset $U$ need to be both informative and diverse
in order to make the best use of the available labelling resources.


\section{Active model checking}
The shape and properties of the functional dependence of satisfaction for a
logical specification with respect to parameters are generally not known a
priori and can exhibit a variety of properties.
For example, in the running example the satisfaction
probability is non-stiff with respect to parameter changes in one direction
while being stiff with respect changes in another.
In addition, much of the sampling was performed in completely flat regions
of the parameter space.
Thus the key challenge addressed in this section is where to sample to make the
posterior estimates as informative as possible about the underlying mechanics.
We aim to decide on the regions where the satisfaction
probability surface is not flat and concentrate most of our model checking
effort there.

In this section we introduce the main contribution of this paper --- active
model checking.
The general outline of the procedure is given by Algorithm~\ref{alg:active}.
\begin{algorithm}
    \caption{Active smoothed model checking}\label{alg:active}
    \begin{algorithmic}[1]
        \Procedure{Model checking}{model $\mathcal{M}$, property $\varphi$, parameter space $\mathcal{X}$}
        \State $\mathcal{D}_{\mathit{old}} 
            \gets \mathit{generate\_initial\_data}(\mathcal{M}, \varphi, \mathcal{X})$
        \State $\mathcal{D}_{\mathit{new}} \gets \mathcal{D}_{\mathit{old}}$
        \State $\vv{u} \gets \mathit{inducing\_points}(\mathcal{D}_{\mathit{old}})$
        \State $\vv{v} \gets \vv{u}$
        \State $q(\vv{v}) \gets \mathit{initialise\_posterior}(\vv{v})$
        \While{$\mathit{true}$}
        \State $q(\vv{v}) \gets 
        \mathit{update\_variational}(q(\vv{u}), \vv{v}, 
        \mathcal{D}_{\mathit{new}}, \mathcal{D}_{\mathit{old}})$
        \State $\mathcal{D}_{\mathit{old}} 
        \gets \mathcal{D}_{\mathit{old}} \cup \mathcal{D}_{\mathit{new}}$
        \State $\mathcal{D}_{\mathit{new}} 
        \gets \mathit{query\_new}(q(\vv{v}), \mathcal{M}, \varphi, \mathcal{X})$
        \State $\vv{u} \gets \vv{v}$
        \State $\vv{v} \gets \mathit{update\_inducing}(\mathcal{D}_{\mathit{new}})$
        \EndWhile
        \State \textbf{return} $q(\vv{v})$
        \EndProcedure
    \end{algorithmic}
\end{algorithm}
The first step, given by the procedure $\mathit{generate\_initial\_data}$,
is to simulate the initial data set $\mathcal{D}_{\mathit{old}}$ via stochastic
simulation of the CTMC model $\mathcal{M}$ for a sample of the
parameter space $\mathcal{X}$ and checking whether or not the individual
trajectories satisfy the property $\varphi$ or not.
The initial set of parameter samples can for example be a regular grid or
sampled uniformly from the parameter space.
We are going to experiment with both initialisations.

Based on the results $\mathcal{D}_{\mathit{old}}$ we choose the inducing points
$\vv{v}$, captured by the procedure $\mathit{inducing\_points}$, $\vv{v}$
based on the approach in Section~\ref{sec:inducing points} and initialise the
variational posterior.
Here we are going to initialise the posterior as a multivariate Gaussian
$q(\vv{v}) = \mathcal{N}(\vv{0}, \vv{I}))$ with $0$ mean and identity covariance
matrix.

Each iteration of the model checking loop will update the variational posterior 
$q(\vv{v})$ and use the fitted approximate posterior to query new points in the
parameter space to perform model checking.
At the end of each iteration, based on the updated data set we update the set of
inducing points.  

There are two issues to be resolved before the procedure can be implemented.
First is that the direct use of ELBO as introduced in Section~\ref{sec:var
inference} does not suffice in the online setting where new data becomes
available in batches. 
Second is the challenge of choosing an appropriate query function that is going
to suggest more points in the parameter space at which to gather more model
checking data.  
These will be addressed in the following sections.

\subsection{Streaming setting}
\label{sec:streaming}
In order to incorporate active learning ideas into the Gaussian process based
model checking approach we need to address the problem that not all of the
training data is available a priori.
For our purposes it is important to be able to conduct inference in a streaming
setting where data is gradually added to the model.  
A naive approach would refit a Gaussian process from scratch every time a
new batch of data arrives.
However, with potentially large data sets this becomes infeasible.
To perform sparse variational inference in a scalable way the
method needs to avoid revisiting previously considered data points.
In particular, we consider the method proposed in~\cite{Bui2017}
that derives a correction to ELBO that allows us to incorporate
streaming data incrementally into the posterior estimate.

The main question is how to update the variational approximation to the
posterior at time step $n$, denoted $q_{\mathit{old}}(\vv{u})$, to form an
approximation at the time step $n+1$, denoted $q_{\mathit{new}}(\vv{v})$.
In the following we note the variational posteriors $q_{\mathit{old}}$ and
$q_{\mathit{new}}$ at $\vv{g}_{\varphi}$ and inducing points $\vv{u}$ and
$\vv{v}$, respectively, are approximations to the true posteriors given
observations $\vv{y}_{\mathit{old}}$ and $\vv{y}_{\mathit{new}}$.
It was shown in~\cite{Bui2017} that the lower bound $\log
p(\vv{y}_{\mathit{new}} |\vv{y}_{\mathit{old}})$ becomes
\begin{flalign*}
    \mathcal{F}(q_{\mathit{new}}(\vv{g}_{\varphi}, \vv{v}))
    = \int q_{\mathit{new}}(\vv{g}_{\varphi}, \vv{v}) 
    \log p(\vv{y}_{\mathit{new}} | \vv{g}_{\varphi}, \vv{v}) d(\vv{g}_{\varphi}, \vv{v}) 
    -\mathcal{D}_{KL}(q_{\mathit{new}}(\vv{v}), p(\vv{v}))
    \\
    - \mathcal{D}_{KL}(q_{\mathit{new}}(\vv{u}), q_{\mathit{old}}(\vv{u}))
    + \mathcal{D}_{KL}(q_{\mathit{new}}(\vv{u}), p(\vv{u}))
\end{flalign*}
The above can be interpreted as follows: the first two terms give the ELBO 
under the assumption that the new data seen at iteration $n+1$ is the whole data
set; the final two terms take into account the old likelihood through the
approximate posteriors at old inducing points and the prior $p(\vv{u})$.
This allows us to implement an online version of the smoothed model checking
where observation data arrives in batches.

\subsection{Query strategies}
As discussed in Section~\ref{sec:active learning}, in order to implement an
active learning method for model checking we need to decide which new
parameters are tested based on the existing information.
In the following we consider two query strategies for active model checking.
\subsubsection{Predictive variance}
The first approach is a commonly used experimental design strategy which aims to
minimise the predictive variance.
Recall that in smoothed model checking for a property $\varphi$ we fit a latent
Gaussian processes $g_{\varphi}$.
The posterior satisfaction probability for parameter $\vv{x}_*$ given the GP
$g_{\varphi}$ is then calculated via
\begin{equation*}
    p(y_* = 1 | \mathcal{D}, \vv{x}_*) = 
    \int \sigma(g_{\varphi}(\vv{x}_*)) 
    p(g_{\varphi}(\vv{x}_*) | \mathcal{D}) 
    dg_{\varphi}(\vv{x}_*)   
\end{equation*}
The above can also be seen as the expectation of $\sigma(g_{\varphi}(\vv{x}_*)$ with
respect to the distribution $g_{\varphi}(\vv{x}_*)$, denoted $\mathbb{E}\left[\sigma(g_{\varphi}(\vv{x}_*) \right] $.
Similarly, we can consider the variance of this estimate
\begin{equation*}
    \mathbb{E}\left[ \sigma(g_{\varphi}(\vv{x}_*)^2 \right] 
    - \mathbb{E}\left[ \sigma(g_{\varphi}(\vv{x}_*) \right]^2
\end{equation*}
Our aim is then to iteratively train the Gaussian process model so that
predictive variance over the parameter space is minimised.

Before giving the outline of the proposed procedure we address the issue of
redundancy in the query points.
As pointed out in Section~\ref{sec:active learning} simply taking a set of top 
points with respect to the utility function, in this case predictive variance,
leads to querying parameters that are clustered together.  
We can overcome this problem by clustering the pool of unlabelled samples
$\mathcal{U}$ from which the query choice is made. 
In particular, the top points with respect to predictive variance are chosen
from a pool of samples where the redundancy is already reduced. 
Informally, this leads to the following basic outline of the procedure:
\begin{enumerate}
    \item Sample an initial set of training points or parametrisations $\vv{x}$
        of the model (via uniform, Latin hypercube sampling or taking points on
        a regular grid) and conduct model checking based on sampled
        trajectories.
        These points are used to choose the set of inducing points via the
        k-means++ clustering algorithm and fit the first iteration of the
        Gaussian process model.
    \item For the next iteration we randomly sample another set of points $U$
        and cluster them via regular kmeans. 
        From the set of cluster centres $U_k$ the query function $\mathcal{Q}$
        selects a set of points for model checking. 
        The query function is simply defined by taking the subset $U^*$ of
        cluster centres $U_k$ where the predictive variance, as defined above,
        is the highest.
        This concentrates the sampling to points where the model is most
        uncertain about its prediction.
    \item The points in $U^*$ are labelled by simulating the model for the
        parametrisations in $U^*$ and checking the resulting trajectories
        against the logic specification $\varphi$.
        The results are incorporated into the Gaussian process model via the
        streaming method discussed in Section~\ref{sec:streaming}.
    \item Repeat points 2 and 3 until a set computational budget is exhausted. 
\end{enumerate}

\subsubsection{Predictive gradient}
The second strategy we consider is based on the predictive mean
$\bar{g}_{\varphi}(\vv{x})$ of the Gaussian process.
Our aim is to concentrate the sampling at the locations where the predictive
mean undergoes the most rapid change.
This requires gradients of the predictive mean.

We recall from Section~\ref{sec:var inference} that for a variational posterior
$q(\vv{u})$ with mean $\bm{\mu}$ and covariance $\bm{\Sigma}$, the posterior
mean at a point $\vv{x}$ is given by
\begin{equation}
    \bar{g}_{\varphi}(\vv{x}) = k(\vv{x}, \vv{u}) K_{mm}^{-1} \bm{\mu} \defas k(\vv{x}, \vv{u}) \bm{\alpha}
\end{equation}
Only the first part, the kernel function, depends on $\vv{x}$.
Thus, in order to get the derivative of the predictive mean we need to
differentiate $k(\vv{x}, \vv{u})$. 
Recall, that in this paper we chose to work with the squared exponential kernel
given by 
\begin{equation}
    k(\vv{x}, \vv{u}_i) = \exp\left( -\frac{|\vv{x} - \vv{u}_i|^2}{2\ell} \right)
\end{equation}
We have used $\vv{u}_i$ to denote a single inducing point in the set of inducing
points $\vv{u}$.
Thus, the derivative of $k(\vv{x}, \vv{u}_i)$  with respect to $\vv{x}$ is given
by 
\begin{equation}
    \frac{dk(\vv{x}, \vv{u}_i)}{d\vv{x}} 
    = 
    - \frac{\vv{x} - \vv{u}_i}{\ell}
    \exp\left( -\frac{|\vv{x} - \vv{u}_i|^2}{2\ell} \right) = 
    - \frac{\vv{x} - \vv{u}_i}{\ell} k(\vv{x}, \vv{u}_i)
\end{equation}
As the above is for a single inducing point $\vv{u}_i$, to compute the
derivative of the posterior mean we need to concatenate this derivative for all
$m$ inducing points.
Thus, we get
\begin{align*}
    \frac{d\bar{g}_{\varphi}(\vv{x})}{d\vv{x}} 
    = 
    -\ell^{-1} 
    \begin{bmatrix}
        \vv{x} - \vv{u}_1 \\
        \cdots \\
        \vv{x} - \vv{u}_m 
    \end{bmatrix}
    (k(\vv{x}, \vv{u}) \odot \bm\alpha)
\end{align*}
where $\odot$ denotes element-wise multiplication.
Given this we can proceed as in the case of the predictive variance.
The only change is that instead of considering the predictive variance
for each sampled set of parameters we calculate the norm of
$\frac{d\bar{g}_{\varphi}(\vv{x})}{d\vv{x}}$ and define the query function to
choose a subset $U^*$ of cluster centres $U_k$ with the highest norms.

\begin{figure}[t!]
    \centering
    \includegraphics[width=0.8\textwidth]{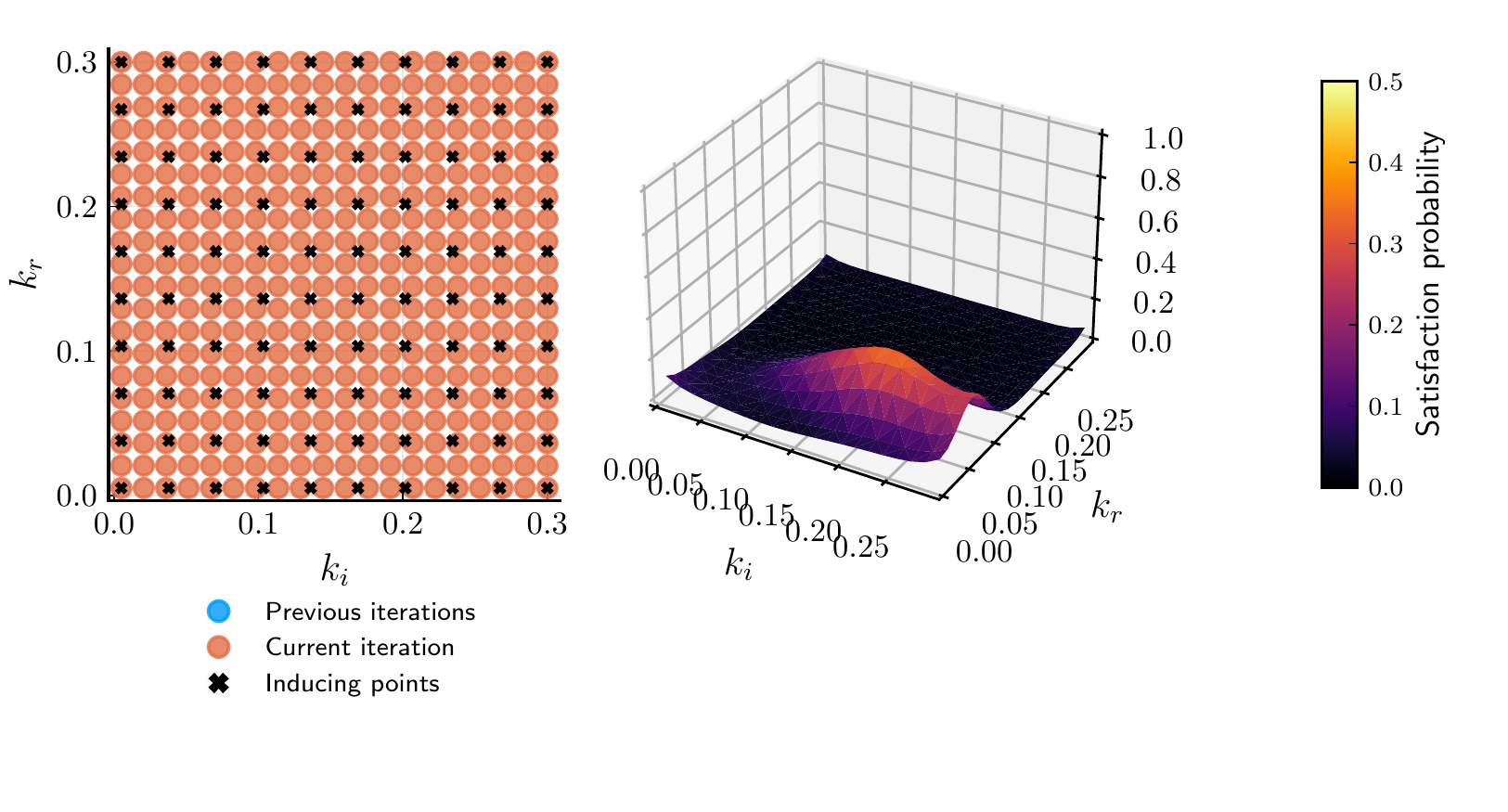}
    \vspace{-3em}
    \caption{Mean satisfaction probability surface. Sparse smoothed MC.}
    \label{fig:surface sparse}
\end{figure}

\subsection{Results}
In this section we evaluate the proposed active learning method for
model checking on the running SIR example.
The methods are compared to the baseline naive stochastic simulation-based
model checking and smoothed model checking without the sparse approximation
and the active step.
We present several metrics for comparing the smoothed model checking results with
the empirical mean based on stochastic simulation.
The first is the mean and standard deviation of the difference between the mean
probability predicted by the fitted Gaussian processes and the empirical mean 
from the stochastic simulation results at each of the points on the $20\times20$
grid.
Secondly, we consider the maximum difference between the predicted mean
probability and the naive empirical mean. 
Finally, we give the root-mean-square error (RMSE)
$\sqrt{\frac{1}{N}\sum{\left(\bar{g}_{\varphi}(\vv{x}_{i}) -
\bar{f}_{\varphi}(\vv{x}_i)\right)^2}}$
where $\bar{g}_{\varphi}(\vv{x}_{i})$ is the predicted mean
satisfaction probability for parametrisation $\vv{x}_i$.
We denote by $\bar{f}_{\varphi}(\vv{x}_i)$ the empirical estimate of the
satisfaction probability at $\vv{x}_i$ given 2000 sample trajectories.

In the active learning experiments we start with a coarser $10\times10$ grid
followed by an active iteration where an additional 300 points are chosen to
refine the approximation for a total of 400 training points.
The inducing points are initialised by choosing the initial grid of $100$ points
as inducing points and kept constant for the remainder of the fitting procedure.
Similarly we present the results for sparse smoothed model checking for a
$20\times20$ grid with the $10\times10$ grid of inducing points as well as
smoothed model checking where inducing points are not chosen.
Parameters of the kernel function are kept constant in each of the smoothed
model checking experiments.

\subsubsection{SIR}

\begin{table}[t!]
    \caption{Comparison of accuracy for smoothed model checking and the sparse
    and active learning extensions. SIR model. Naive threshold greater than 0.02.} 
    \label{tab:smoothed mc vs active acc}
    \centering
    \footnotesize
    \begin{tabular}[t]{@{\extracolsep{4pt}}llcc@{}}
    \toprule
        Method & error mean/var & maximum & RMSE \\
    \midrule
        Smoothed MC & $(0.042, 0.032)$ & $0.13$ & $0.578$ \\
        Sparse smoothed MC & $(0.049, 0.036)$ & $0.151$ & $0.664$ \\
        Active sparse smoothed MC &  &  &  \\
            \qquad Predictive variance & $(0.044, 0.038)$ & $0.16$ & 0.629 \\
            \qquad Predictive gradient & $\mathbf{(0.044, 0.033)}$ & $\mathbf{0.136}$ & $\mathbf{0.6}$ \\
            \qquad Random sampling     & $(0.049, 0.04)$ & 0.21 & 0.684 \\
    \bottomrule
    \end{tabular}
\end{table}

\begin{table}[t!]
    \caption{Comparison of computation times for smoothed model checking and the
    sparse and active learning extensions. SIR model.} 
    \label{tab:smoothed mc vs active time}
    \centering
    \footnotesize
    \begin{tabular}[t]{@{\extracolsep{4pt}}lcccc@{}}
    \toprule
        Method & SSA & Inference & Query & Total \\
    \midrule
        Naive statistical MC & 59.5 & N/A & N/A & 59.5\\
        Smoothed MC & 1.6 & 24.6 & N/A & 26.2 \\ 
        Sparse smoothed MC & 1.6 & 4.6 & N/A & $\mathbf{6.2}$ \\ 
        Active sparse smoothed MC & & &\\ 
            \qquad Predictive variance & 1.7 & 6.2 & 1.9 & 8.6 \\ 
            \qquad Predictive gradient & 1.8 & 5.0 & 0.4 & 8.5 \\ 
            \qquad Random sampling     & 1.6 & 5.7 & 0.00 & 7.3 \\ 
    \bottomrule
    \end{tabular}





\end{table}

We again consider the SIR model.
The results for accuracy are summarised in Table~\ref{tab:smoothed mc vs
active acc}.
Table~\ref{tab:smoothed mc vs active acc} gives the comparisons for each point
on the $20\times20$ grid where the naive model checking was conducted.
We restrict the view to those points where the naive model checking-based
satisfaction probability estimates exceed $0.02$ in order to consider locations
where the surface is not completely flat.
Note that we would expect the full smoothed MC without inducing points to offer
the best accuracy due to the fact that fewer approximations are made in the
inference algorithm.
However, the active methods with predictive variance and gradient-based query
functions provide a better approximation than the sparse model checking without
an active step.
The benefit of the sparse methods comes from significant reductions in
computation costs.
From Table~\ref{tab:smoothed mc vs active time} we see that there is some
overhead associated with implementing the active learning procedures compared to
simply exhausting the chosen computational budget of 400 points.
However, as expected the sparse methods offer a significant speed-up compared to
the smoothed model checking with no inducing points.

\begin{figure}[t!]
    \centering
    \begin{subfigure}[t]{0.8\textwidth}
        \includegraphics[width=\textwidth]{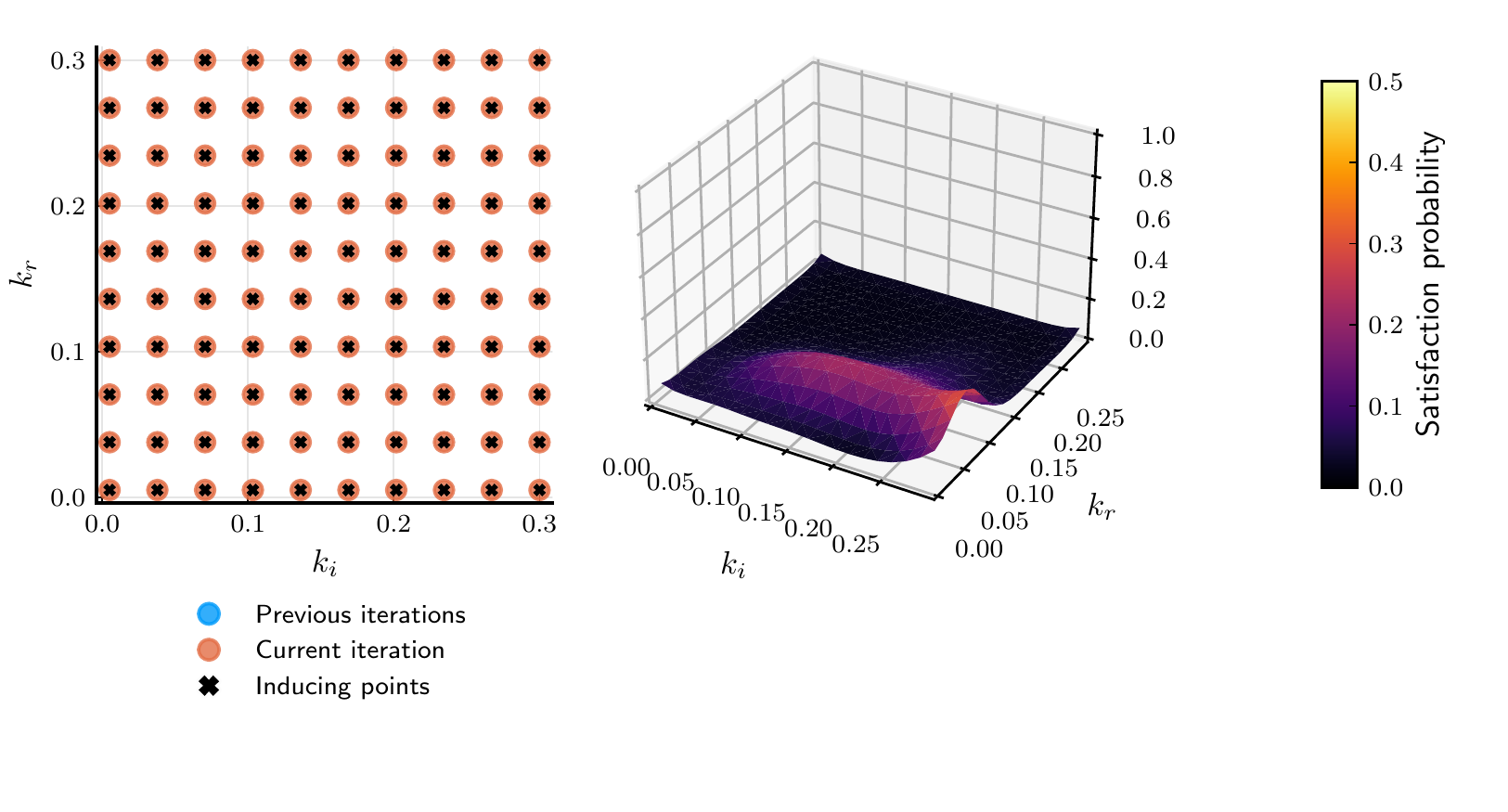}
    \end{subfigure}

    \vspace{-2.0em}
    \begin{subfigure}[t]{0.8\textwidth}
        \includegraphics[width=\textwidth]{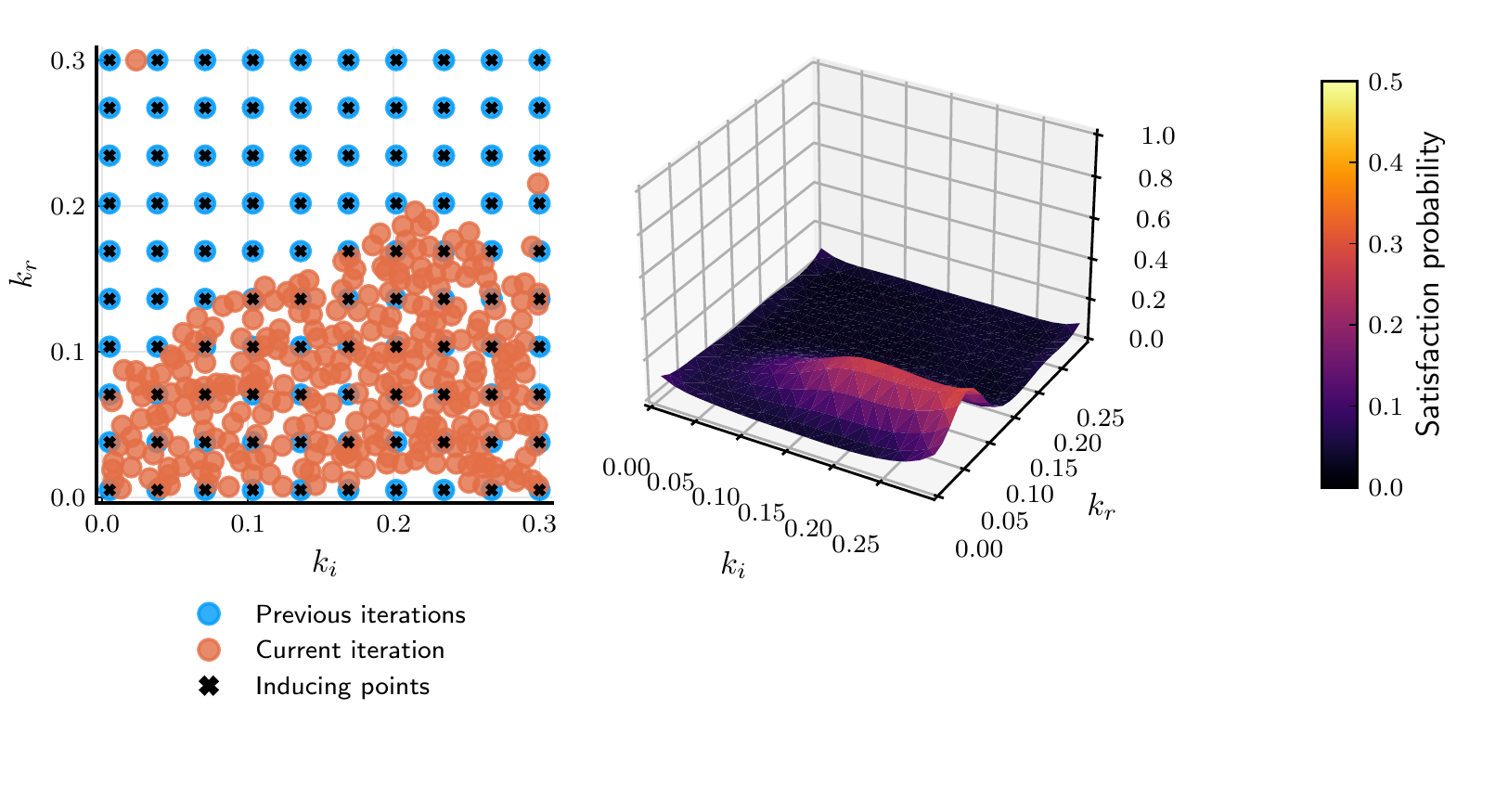}
    \end{subfigure}
    \vspace{-1.5em}
    \caption{Mean satisfaction probability surface. Active and sparse smoothed
        model checking with 2 iterations. Predictive variance-based query function.}
    \label{fig:surface predvar}
\end{figure}

\begin{figure}[t!]
    \centering
    \begin{subfigure}[t]{0.8\textwidth}
        \includegraphics[width=\textwidth]{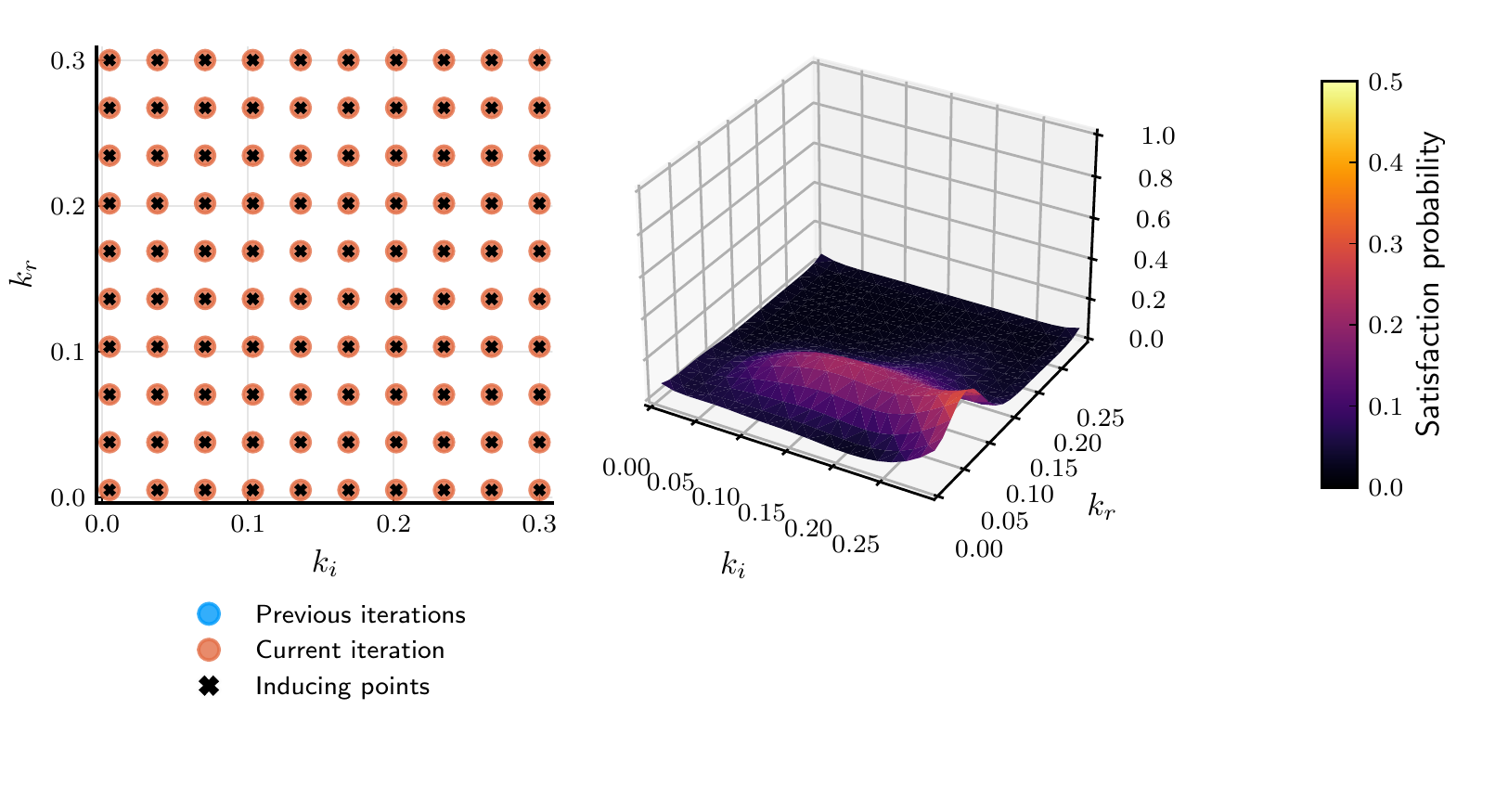}
    \end{subfigure}

    \vspace{-2.0em}
    \begin{subfigure}[t]{0.8\textwidth}
        \includegraphics[width=\textwidth]{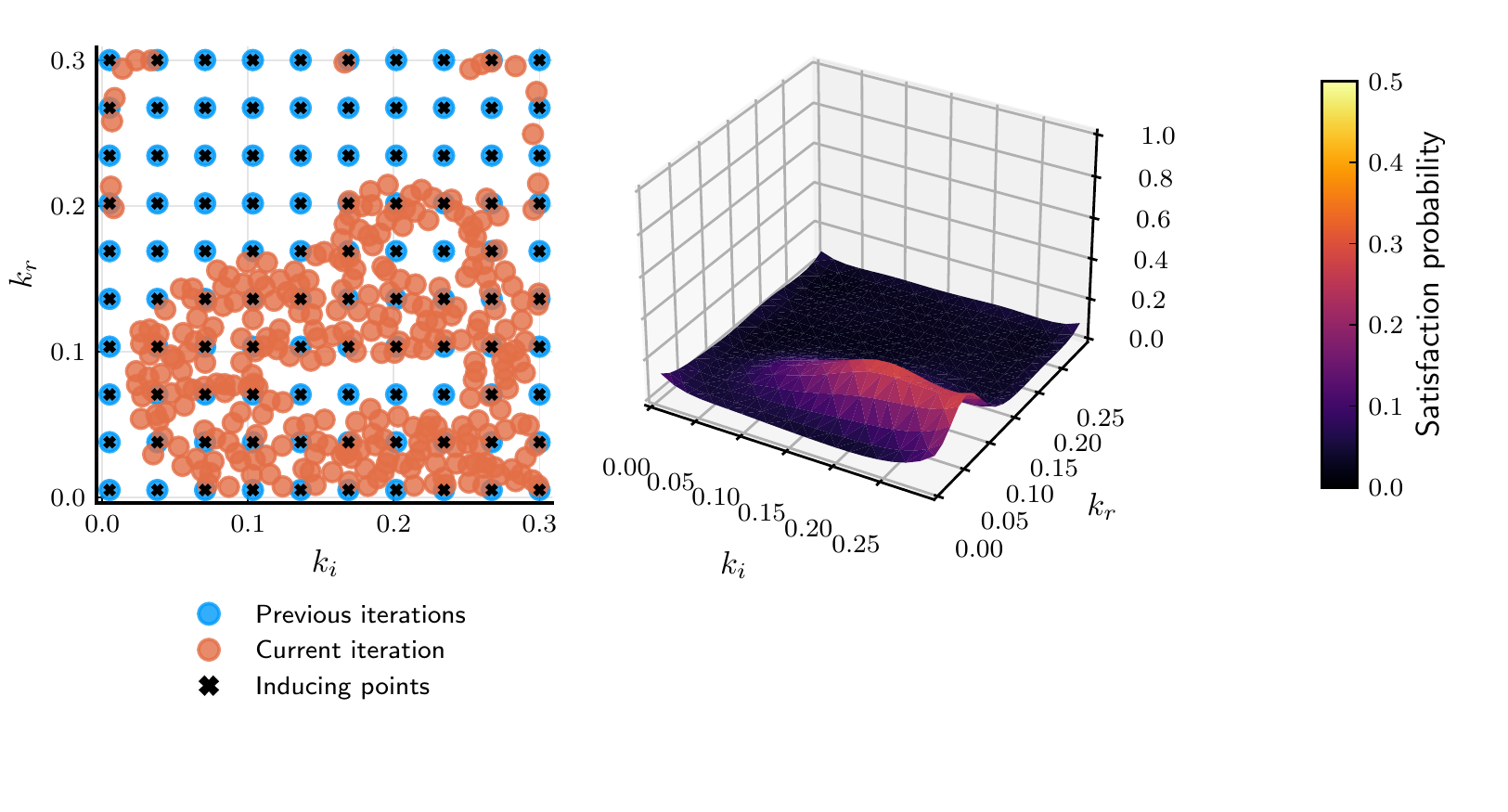}
    \end{subfigure}
    \vspace{-1.5em}
    \caption{Mean satisfaction probability surface. Active and sparse smoothed
        model checking with 2 iterations. Predictive mean gradient-based query function.}
    \label{fig:surface predgrad}
\end{figure}

Figure~\ref{fig:surface sparse} gives the mean satisfaction probability surface
based on the fitted sparse Gaussian process. 
Figures~\ref{fig:surface predvar} and ~\ref{fig:surface predgrad} present the
evolution of the predictive mean surface through two active learning
iterations.
All figures are accompanied by the scatter plots showing where the samples were drawn.


%
%
%
\subsection{Implementation}
The prototype implementation in written in the Julia programming language and
makes use of the tools provided as part of Julia Gaussian Processes
repositories~\cite{JuliaGaussian} to set up the Gaussian process models.
The CTMC models are defined as chemical reaction networks with tools provided as
part of the SciML ecosystem for scientific simulations~\cite{Rackauckas2017}.
The simulations were carried out on a laptop with Intel i7-10750H CPU.

\section{Conclusions}
In this paper we applied sparse approximation and active learning to smoothed
model checking.
By leveraging existing sparse approximations, we improved scalability of the
inference algorithms for Gaussian process classification corresponding to the
smoothed model checking problem.
Additionally, we showed that by concentrating the sampling to high variance or  
high predictive gradient areas of the parameter space, we improved the resulting
approximation compared to sparse models with uniform or grid-based sampling
of model parameters.
When compared to the standard smoothed model checking approach with no inducing
point approximation and no active step, our method significantly speeds up the
inference procedure while attempting to reduce errors inherent in sparse
approximations.
This aligns with the pre-existing results from active learning literature which 
aim to construct learning algorithms that actively query for observations in
order to improve accuracy while keeping the number of observations needed to
a minimum.  

As further work, we aim to refine our query methods and make a comparison
with other existing methods in the active learning literature.
Secondly, we plan to link the choice of inducing points to the active query
methods more directly.
In particular, we will test if the inducing points, and perhaps the
underlying kernel parameters, can be effectively reconfigured through active
iterations.
This would further improve the approximation to the satisfaction probability
surface at the non-stiff parts of the parameter space.
Finally we will consider alternative kernel functions. 
The kernel function chosen in this paper is a standard first approach in many
settings but is best suited for modelling very smooth functions --- not
necessarily the case with satisfaction probability surfaces for parametric
CTMCs. 

\bibliographystyle{splncs04}
\bibliography{bibliography}

\begin{thebibliography}{10}
\providecommand{\url}[1]{\texttt{#1}}
\providecommand{\urlprefix}{URL }
\providecommand{\doi}[1]{https://doi.org/#1}

\bibitem{JuliaGaussian}
Gaussian processes for machine learning in julia.
  \url{https://github.com/JuliaGaussianProcesses}, accessed: 2021-01-14

\bibitem{AghaP18}
Agha, G., Palmskog, K.: A survey of statistical model checking. {ACM} Trans.
  Model. Comput. Simul.  \textbf{28}(1),  6:1--6:39 (2018)

\bibitem{David2007}
Arthur, D., Vassilvitskii, S.: K-means++: The advantages of careful seeding.
  In: Proceedings of the Eighteenth Annual ACM-SIAM Symposium on Discrete
  Algorithms. p. 1027–1035. SODA, Society for Industrial and Applied
  Mathematics, USA (2007)

\bibitem{BortolussiMS16}
Bortolussi, L., Milios, D., Sanguinetti, G.: Smoothed model checking for
  uncertain continuous-time markov chains. Inf. Comput.  \textbf{247},
  235--253 (2016)

\bibitem{BortolussiS18}
Bortolussi, L., Silvetti, S.: Bayesian statistical parameter synthesis for
  linear temporal properties of stochastic models. In: {TACAS} 2018,. Lecture
  Notes in Computer Science, vol. 10806, pp. 396--413. Springer (2018)

\bibitem{Bui2017}
Bui, T.D., Nguyen, C.V., Turner, R.E.: Streaming sparse gaussian process
  approximations. In: Advances in Neural Information Processing Systems 30:
  Annual Conference on Neural Information Processing Systems 2017. pp.
  3299--3307 (2017)

\bibitem{Csato2002}
Csato, L., Opper, M.: Sparse online gaussian processes. Neural Computation
  \textbf{14},  641--668 (2002)

\bibitem{DonzeM10}
Donz{\'{e}}, A., Maler, O.: Robust satisfaction of temporal logic over
  real-valued signals. In: {FORMATS} 2010. Proceedings. Lecture Notes in
  Computer Science, vol.~6246, pp. 92--106. Springer (2010)

\bibitem{Hernandez-Lobato2016}
Hernandez-Lobato, D., Hernandez-Lobato, J.M.: Scalable gaussian process
  classification via expectation propagation. In: Gretton, A., Robert, C.C.
  (eds.) Proceedings of the 19th International Conference on Artificial
  Intelligence and Statistics. Proceedings of Machine Learning Research,
  vol.~51, pp. 168--176. PMLR (2016)

\bibitem{Jha2009}
Jha, S.K., Clarke, E.M., Langmead, C.J., Legay, A., Platzer, A., Zuliani, P.: A
  bayesian approach to model checking biological systems. In: Degano, P.,
  Gorrieri, R. (eds.) Computational Methods in Systems Biology, 7th
  International Conference, {CMSB}. Lecture Notes in Computer Science,
  vol.~5688, pp. 218--234. Springer (2009). \doi{10.1007/978-3-642-03845-7\_15}

\bibitem{KwiatkowskaNP07}
Kwiatkowska, M.Z., Norman, G., Parker, D.: Stochastic model checking. In:
  {SFM}. Lecture Notes in Computer Science, vol.~4486, pp. 220--270. Springer
  (2007)

\bibitem{Legay2019}
Legay, A., Lukina, A., Traonouez, L., Yang, J., Smolka, S.A., Grosu, R.:
  Statistical model checking. In: Steffen, B., Woeginger, G.J. (eds.) Computing
  and Software Science - State of the Art and Perspectives, Lecture Notes in
  Computer Science, vol. 10000, pp. 478--504. Springer (2019).
  \doi{10.1007/978-3-319-91908-9\_23}

\bibitem{MalerN04}
Maler, O., Nickovic, D.: Monitoring temporal properties of continuous signals.
  In: FORMATS/FTRTFT 2004, Grenoble, France, September 22-24, Proceedings.
  Lecture Notes in Computer Science, vol.~3253, pp. 152--166. Springer (2004)

\bibitem{Minka2001}
Minka, T.P.: Expectation propagation for approximate bayesian inference. In:
  Breese, J.S., Koller, D. (eds.) {UAI}: Proceedings of the 17th Conference in
  Uncertainty in Artificial Intelligence. pp. 362--369. Morgan Kaufmann (2001)

\bibitem{Rackauckas2017}
Rackauckas, C., Nie, Q.: {Differentialequations.jl}--a performant and
  feature-rich ecosystem for solving differential equations in julia. Journal
  of Open Research Software  \textbf{5}(1) (2017)

\bibitem{Rasmussen2006}
Rasmussen, C.E., Williams, C.K.I.: Gaussian processes for machine learning.
  Adaptive computation and machine learning, {MIT} Press (2006)

\bibitem{Santner2003}
Santner, T.J., Williams, B.J., Notz, W.I.: The Design and Analysis of Computer
  Experiments. Springer series in statistics, Springer (2003)

\bibitem{Sen2004}
Sen, K., Viswanathan, M., Agha, G.: Statistical model checking of black-box
  probabilistic systems. In: Alur, R., Peled, D.A. (eds.) Computer Aided
  Verification, 16th International Conference, {CAV}. Lecture Notes in Computer
  Science, vol.~3114, pp. 202--215. Springer (2004).
  \doi{10.1007/978-3-540-27813-9\_16}

\bibitem{Settles2011}
Settles, B.: From theories to queries. In: Active Learning and Experimental
  Design workshop, In conjunction with {AISTATS}. {JMLR} Proceedings, vol.~16,
  pp. 1--18. JMLR.org (2011),
  \url{http://proceedings.mlr.press/v16/settles11a/settles11a.pdf}

\bibitem{Settles2012}
Settles, B.: Active Learning. Synthesis Lectures on Artificial Intelligence and
  Machine Learning, Morgan {\&} Claypool Publishers (2012).
  \doi{10.2200/S00429ED1V01Y201207AIM018}

\bibitem{Titsias2009}
Titsias, M.K.: Variational learning of inducing variables in sparse gaussian
  processes. In: Dyk, D.A.V., Welling, M. (eds.) Proceedings of the Twelfth
  International Conference on Artificial Intelligence and Statistics,
  {AISTATS}. {JMLR} Proceedings, vol.~5, pp. 567--574 (2009)

\end{thebibliography}

\end{document}